\documentclass[10pt, a4paper]{article}
\usepackage{lrec}
\usepackage{amsmath}
\usepackage{multibib}
\usepackage{enumitem}
\newcites{languageresource}{Language Resources}
\usepackage{graphicx}
\usepackage{tabularx}
\usepackage{soul}
\usepackage{booktabs}
\usepackage{todonotes} 
\usepackage{epstopdf}
\usepackage[latin1]{inputenc}

\usepackage{hyperref}
\usepackage{xstring}

\usepackage{latexsym}
\usepackage{url}
\usepackage{enumitem}

\usepackage{graphicx}
\usepackage{multirow}
\usepackage[linesnumbered,ruled,vlined]{algorithm2e}
\usepackage{todonotes}
\usepackage{times}

\SetAlFnt{\small}
\SetAlCapFnt{\small}
\SetAlCapNameFnt{\small}
\SetAlCapHSkip{0pt}
\IncMargin{-\parindent}
\usepackage{xcolor}% http://ctan.org/pkg/xcolor

\SetCommentSty{mycommfont}
\usepackage{todonotes}

\newcommand{\nosemic}{\renewcommand{\@endalgocfline}{\relax}}% Drop semi-colon ;
\newcommand{\dosemic}{\renewcommand{\@endalgocfline}{\algocf@endline}}% Reinstate semi-colon ;
\let\oldnl\nl% Store \nl in \oldnl
\newcommand{\nonl}{\renewcommand{\nl}{\let\nl\oldnl}}% Remove line number for one line

\newcommand{\word}[1]{\emph{#1}}            % mention of word
\newcommand{\sense}[1]{\emph{#1}}             % mention of sense
\usepackage{graphicx}
\usepackage{multirow}
\usepackage{url}
\usepackage{listings}

\usepackage{booktabs} % For formal tables

\title{Enriching Frame Representations with Distributionally Induced Senses}

\name{Stefano Faralli$^\dag$, Alexander Panchenko$^\ddag$, Chris Biemann$^\ddag$, Simone Paolo Ponzetto$^\dag$}

\address{$^\dag$Data and Web Science Group, School of Business Informatics and Mathematics, Universit\"{a}t Mannheim, Germany \\ $^\ddag$ Language Technology Group, Department of Informatics, Universit\"{a}t Hamburg, Germany\\
		 \url{{stefano,simone}@informatik.uni-mannheim.de} \\
         \url{{panchenko,biemann}@informatik.uni-hamburg.de} \\ }
\abstract{
We introduce a new lexical resource that enriches the Framester knowledge graph, which links Framnet, WordNet, VerbNet and other resources, with semantic features from text corpora. These features are extracted from distributionally induced sense inventories and subsequently linked to the manually-constructed frame representations to boost the performance of frame disambiguation in context. Since Framester is a frame-based knowledge graph, which enables full-fledged OWL querying and reasoning, our resource paves the way for the development of novel, deeper semantic-aware applications that could benefit from the combination of knowledge from text and complex symbolic representations of events and participants. Together with the resource we also provide the software we developed for the evaluation in the task of Word Frame Disambiguation (WFD).
\newline 
}

\begin{document}

\maketitleabstract

\let\OldUrlFont\UrlFont
\renewcommand{\UrlFont}{\scriptsize\OldUrlFont}

\section{Introduction}

Recent years have witnessed an impressive amount of work on the automatic construction of wide-coverage knowledge resources. Web-scale information extraction systems like NELL \cite{carlson10} or Knowledge Vault \cite{Dong14} can acquire massive amounts of machine-readable knowledge from the Web, whereas projects like DBpedia \cite{bizer09}, YAGO \cite{rabele16} or BabelNet \cite{navigli12} have turned collaboratively-generated content into large knowledge bases. 
However, all of these resources are entity-centric in that they are primarily built around the notion of \emph{entities}, as either provided by an external resource (e.g., Wikipedia pages) or automatically discovered from text (e.g., by clustering entity mentions). The entities are most commonly represented by nouns, noun phrases, and named entities. Entities lie at the heart of the Semantic Web and are central to enable semantic technologies on a large scale \cite{dietz17}. Besides, they also provide the foundation for more complex semantic representations like event templates or frames \cite{fillmore:case}, which are in the focus of our interest in this paper.

Recent work looked at ways to populate the Linked Open Data (LOD) cloud with wide-coverage information about semantic frames on the basis of Framester \cite{gangemi16}, a frame-centric resource that is meant to act as a hub between several other linguistic resources that are already part of the LOD cloud. In parallel, researchers looked at ways to combine knowledge graphs of this kind with distributional semantics to include human-readable meaning representations based on semantic vector spaces within the Semantic Web ecosystem \cite{Farallietal:2016}. Consequently, in this paper we bring these two lines of research together and present a new resource that combines a wide-coverage symbolic repository of frames with human-readable distributional semantic representations from text. 

%\begin{table*}[t!]
%\centering
%\scalebox{0.8}{
%\begin{tabular}{c|ll|ll|ll|ll|}
%\multicolumn{7}{l}{\textbf{\normalsize PCZ (wiki-n200-380k)}}\\
%\hline                                                         
%\multirow{7}{*}{\rotatebox[origin=c]{90}{skos:related}}&\multicolumn{2}{l}{read\#VB\#1}&\multicolumn{2}{|l|}{reader\#NN\#0}&\multicolumn{2}{|l|}{publication\#NN\#2}&\multicolumn{2}{|l|}{reread\#VB\#0}\\
%\hline                            
%& quote\#VB\#0 & 1.0 &viewer\#NN\#0& 1.0 & magazine\#NN\#0 & 1.0&read\#VB\#1&1.0\\\
%& recite\#VB\#0 & .79 & audience\#NN\#1&.79&journal\#NN\#1 & .79&interpret\#VB\#0&.79\\
%& understand\#VB\#0 & .69 & listener\#NN\#1&.69&periodical\#NN\#0 & .69&write\#VB\#0&.69\\
%& interpret\#VB\#0 & .63 & public\#NN\#0&.63&newspaper\#NN\#0 & .63& copy\#VB\#0&.69 \\
%& hear\#VB\#0 & .58 & fan\#NN\#0&.58&papers\#NN\#0 & .58 & rewrite\#VB\#0&.58 \\
%& (190+ more ...) &  & (190+ more ...)& & (140+ more ...) &  & (190+ more ...) &\\
%\hline                                      
%\end{tabular}}
%\caption{An example of four distributional-based senses from a proto-conceptualization.\label{tab:runningddt}}
%\end{table*}

\begin{figure*}[t]
 \centering
 \includegraphics[width=.9\textwidth]{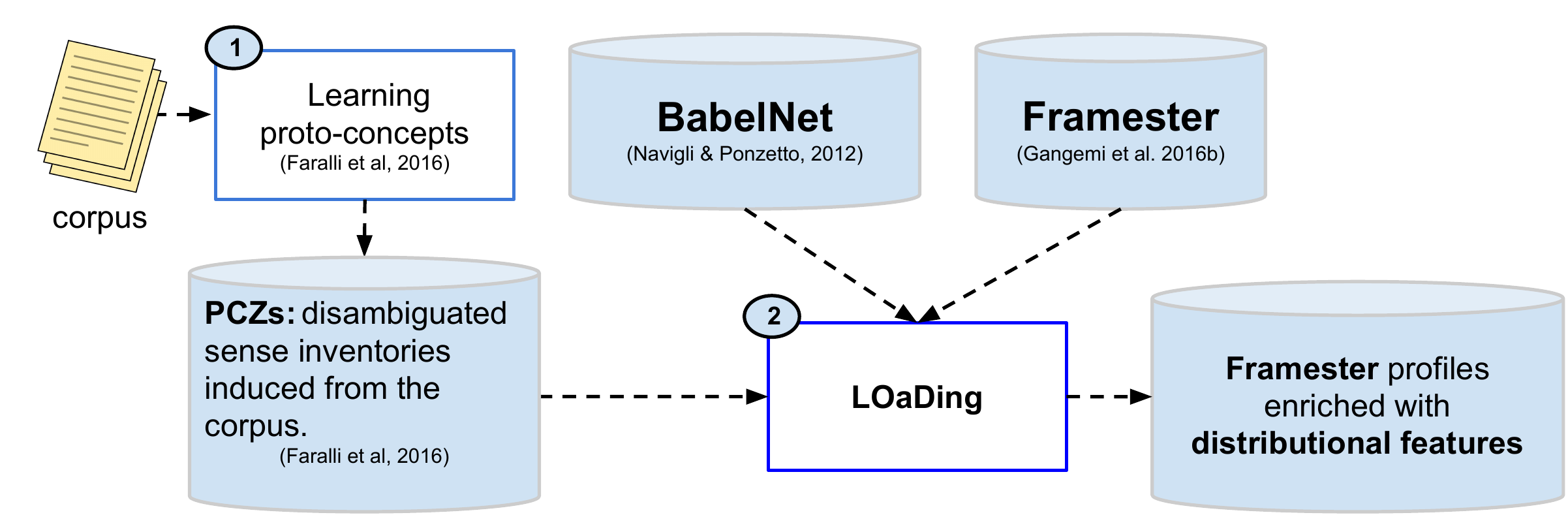}
 \caption{The overall resource construction workflow of our approach: distributionally induced sense representations extracted from a corpus (PCZ) are used to enrich the Framester symbolic knowledge graph.   \label{fig:workflow}}
\end{figure*}

In Table \ref{tab:runningddt}, we show an excerpt from a distributional-based sense inventory, a proto-conceptualization (PCZ) proposed by \newcite{Farallietal:2016} and further developed by~\newcite{biemann2018}. The senses in this resource are induced automatically from text via clustering of graphs of semantically related words and linked to BabelNet and WordNet lexical resources. In particular, in the resource, each sense (e.g., \sense{read\#VB\#1})
%\footnote{In the following, we use \word{Sans Serif} for words and queries, \term{italics} for vocabulary terms and sense inventory items, and \framesem{Small Caps} for frame labels.}
is defined by its related senses (\sense{quote\#VB\#0}, \sense{recite\#VB\#0}, and so on) which have been observed to co-occur with the sense.  Sense inventories with related senses as well as other features, such as weighted hypernymy senses and context clues, can provide rich and human-readable knowledge to disambiguate word meanings in context, e.g., the meaning for the verb \word{to read} as `read aloud' versus `make sense' as in:
\vspace{1em}
\begin{flushleft}
When we \textit{hear} someone \textsc{read} a text, our \textit{understanding} of what we \textit{hear} is usually spontaneous -- the rules by which we \textit{interpret} meaning. \hfill
\end{flushleft}
\vspace{1em}
Here, the reference to the `Reading\_aloud' frame from FrameNet\footnote{\url{https://framenet.icsi.berkeley.edu/}} could be triggered on the basis of the occurrences of the verbs in the sentence: \word{hear}, \word{understanding} and \word{interpret}. Such connections, in turn, could be provided by a hybrid resource where distributional representations from text have been explicitly linked to semantic knowledge repositories, since hybrid resources of this kind have been shown in the past to improve performance on lexical understanding \cite{panchenko2017sense} as well as taxonomy learning and cleaning \cite{Farallietal:2017}.

% Framester links FrameNet, WordNet, VerbNet, BabelNet, DBpedia, Yago, and DOLCE-Zero lexical resources creating a connected knowledge graph. The formal specification in OWL allows querying and reasoning on this linked resource.
% Besides, it applies a formal treatment for Fillmore's frame semantics, enabling OWL querying and reasoning on the linked resource.

In this paper, we bridge the gap between distributional and frame semantics by linking distributional semantic representations to Framester, a knowledge graph that acts as a hub between resources like FrameNet, BabelNet and DBpedia, among others. As a result of this, we introduce a new lexical resource that enriches the Framester knowledge graph with distributional features extracted from text, and show how this hybrid resource yields better results on the task of recognizing frames in running text.

% These have the potential to better recognize frames in running text. 

\begin{table*}[t]
\centering

\begin{tabular}{l|p{4.5cm}|p{8.5cm}}
\textbf{BabelNet Synset ID} & \textbf{English Word Senses} & \textbf{Gloss} \\
\toprule
bn:00085007v & quote, cite & `To repeat someone's exact words.', `To quote; to repeat, as a passage from a book, or the words of another.'\\
\midrule
bn:00090740v & observe, mention, note, remark & `She observed that his presentation took up too much time.',  `They noted that it was a fine day to go sailing.'\\
\end{tabular}
\caption{Excerpt of two BabelNet synsets.\label{tab:runningbn}}
\end{table*}

\begin{table}[t]
\centering
\begin{tabular}{l|l}
\textbf{Frame ID} & \textbf{Related BabelNet Synsets}\\
\toprule
Evidence &  bn:00085007v, bn:00084633v, \dots \\
Telling &   bn:00085007v, bn:00083488v, \dots \\ 
Communication & bn:00085007v, bn:00090740v, \dots\\ 
\end{tabular}
\caption{Excerpt of the Framester TransX profile.\label{tab:runningframester}}
\end{table}

Joining distributional and frame semantics builds upon and extends our framework for combining symbolic and statistical meaning representations~\cite{biemann2018}: the main objective of this line of research is to `join forces' across heterogeneous knowledge and semantic models in order to mutually enrich them and combine the strengths of the lexicographic tradition that describes linguistic information manually with the coverage and versatility of the corpus and data-driven approaches that derive meaning representations directly from the data. Bringing together the `best of both worlds' has the potential to combine the benefits of wide-coverage symbolic \cite{gangemi16b} and statistical \cite{7078639} semantics for frame parsing, which, in turn, can be exploited for many different applications ranging from sentiment analysis \cite{recupero15} all the way to content-based recommendations \cite{declercq14b}.

\vspace{1em}
The contributions of this paper are the following ones:
\begin{enumerate}[leftmargin=4mm]
\item We present the \textit{LOaDing} lexical resource, an extension of Framester that adds distributional-based features \cite{Farallietal:2016,biemann2018} to frame representations (see Section \ref{sec:loading});
\item We evaluate the resource in the task of Word Frame Disambiguation (WFD), i.e., on the identification of frames in context  \cite{gangemi16b}, and show significant improvements over the original Framester frame representations (see Section \ref{sec:WFD}).
\end{enumerate}
Both resources and software produced in this work are available under a CC-BY 4.0 license.\footnote{\begin{small}\url{http://web.informatik.uni-mannheim.de/joint/}\end{small}}

\begin{table*}[t]
\centering
\begin{tabular}{l|l|p{5.8cm}|p{4.5cm}}
\textbf{PCZ Sense ID} & \textbf{BabelNet Synset ID} & \textbf{Related Senses} & \textbf{Context Clues}\\
\toprule
quote\#VB\#0 & bn:00085007v &cite\#NN\#1$[1.0]$, interview\#VB\#0$[0.8]$, mention\#VB\#1$[0.7]$, publish\#VB\#0$[0.6]$, review\#VB\#0$[0.6]$, \dots &  in\#IN\#pcomp$[24799.3]$,  Register\#NP\#prep\_on$[21282.9]$, Track\#NP\#-vmod$[16808.9]$, \dots \\
\midrule
mention\#VB\#1 & bn:00090740v & attest\#VB\#0$[1.0]$, describe\#VB\#0$[0.8]$, document\#VB\#0$[0.7]$, quote\#VB\#0$[0.6]$, \dots & 
Register\#NP\#prep\_on$[45477.7]$, Track\#NP\#-vmod$[35850.4]$,
say\#VB\#prepc\_as$[17041.8]$, \dots \\
\end{tabular} 
\caption{Excerpt from a proto-conceptualization (PCZ) of two distributional-based senses. Numbers behind related senses are normalized similarity scores, numbers behind context clues are association scores.\label{tab:runningddt}}
\end{table*}

\begin{table*}[t]
\centering

\begin{tabular}{l|l|l}
\textbf{Frame ID} & \textbf{Related BabelNet Synsets} & \textbf{Related PCZ Senses} \\
\toprule

Evidence &  bn:00084633v$[29.0]$, bn:00085007v$[13.0]$, \dots & quote\#VB\#0, \dots \\
Telling &   bn:00085007v$[12.0]$, bn:00083488v$[8.0]$, \dots & quote\#VB\#0, \dots  \\ 
Communication & bn:00090740v$[18.0]$, bn:00085007v$[15.0]$, \dots & mention\#VB\#1, quote\#VB\#0, \dots \\ 
\end{tabular}
\caption{Excerpt of a \textit{LOaDing} TransX profile. Numbers behind related synsets $bs$ are the distributional semantics-based weights $w(F,bs)$ we assign in Step 2.4 of our LOaDing approach. \label{tab:runningloading}}
\end{table*}

\section{Enriching Framester with Symbolic Distributional Sense Representations} 
\label{sec:loading}

In Figure \ref{fig:workflow} we show the diagram describing the workflow used to add distributional features to the Framester profiles to enable a better word frame disambiguation. 

\subsection{Resources and Datasets}

Our approach makes use of three linguistic resources: 

\vspace{1em}
\noindent
\textbf{BabelNet} \cite{navigli12}: a multilingual encyclopedic dictionary that connects concepts and named entities in a very large network of semantic relations (see Table \ref{tab:runningbn}).

\vspace{1em}
\noindent
\textbf{Framester} \cite{gangemi16}: a linguistic LOD hub that provides links from FrameNet's frames to semantically related BabelNet senses  (see Table \ref{tab:runningframester}). The current version consists of six different profiles: \textit{Base} contains only manually curated links, whereas the other five (\textit{DirectX}, \textit{FrameBase}, \textit{Fprofile}, \textit{TransX} and \textit{XWFN}) are automatically built extensions or subsets of the \textit{Base} profile.

\vspace{1em}
\noindent

\textbf{Proto-Conceptualizations} (PCZs) \cite{Farallietal:2016}: a fully disambiguated sense inventory automatically induced from text, providing human-readable distributional semantic representations. In this resource, each sense is provided with (see Table \ref{tab:runningddt}): a) a weighted list of semantically-related and hypernymy senses; b) links to existing knowledge bases (i.e., BabelNet); c) context clues used to disambiguate the senses in the context. 

\vspace{1em}
\noindent

\subsection{Combination of Resources}

The goal of our method is a combination of the three resources mentioned in the previous section as illustrated in Table~\ref{tab:runningloading}.
% We said this 50 times already...
% The motivation of such combination is the fact that an extension of FrameNet with distributional features may enable more advanced applications relying on frames.
This is made possible by using BabelNet as a pivot, since Framester provides links from FrameNet to BabelNet (Table \ref{tab:runningframester}) while our PCZs are also linked to BabelNet (Table \ref{tab:runningddt}). As a first result of such a combination we created \textit{LOaDing}, an extension of Framester's profiles where we provide weights for each frame-related BabelNet synset. We show in Table \ref{tab:runningloading} an excerpt of the resulting extended Framester's TransX profile shown in Table \ref{tab:runningframester}.\\
   
The enriched frame representations are built in two main steps as illustrated in Figure \ref{fig:workflow}: 

\vspace{1em}
\noindent
\textbf{Step 1: Learning a proto-conceptualization (PCZ)}. We apply the  methodology from \newcite{biemann2018}, to produce disambiguated sense inventories with entries for nouns and verbs.

\vspace{1em}
\noindent
\textbf{Step 2: LOaDing a PCZ into  Framester}. For each frame $F$, e.g.\ $\textit{Communication}$, and related BabelNet sysnet $bs$, e.g.\ \textit{bn:00085007v}, we assign a weight $w(F,bs)$. In order to compute the weight, we create three bag-of-word representations for each of the three resources listed in the previous section:

\begin{enumerate}[leftmargin=4mm]

\item $B_{bn}=BoW(bs)$ to represent $bs$ including the counts for all the content words from the synset word senses and the glosses;

\item From the PCZ we collect all the entries $PCZ(bs)$ matching $bs$, for instance $PCZ($\textit{bn:00085007v}$)=\{$\textit{quote\#VB\#0}$,\dots\}$, and create a bag of words $B_z=BoW(PCZ(bs))$ by collecting all the weights of content words from the senses $s \in PCZ(bs)$ and the corresponding related senses.

\item We create a BoW from $B_f$ by collecting all the content words from the description of the frame $F$. 

\item Finally, we compute a relatedness score between a frame $F$ and a BabelNet synset $bs$:
\begin{equation*}
w(F,bs)=\sum_{w \in I}{B_{z}.c(w)\times (B_{f}(F).c(w)+B_{bn}.c(w))},
\end{equation*}
where $I=B_z(bs) \cap (B_{bn}(bs) \cup B_{f}(F))$ and $A.c(w)$ equals the number of occurrences of the word $w$ in $A$. For instance, with respect to the excerpts of Tables \ref{tab:runningddt} and \ref{tab:runningloading} we obtain $w($\textit{Communication}, \textit{bn:00085007v}$)=15.0$.
\end{enumerate}

\section{Using the Enriched Representations for Word Frame Disambiguation}
\label{sec:WFD}

We evaluate our extensions of Framester profiles following the experimental setting of \newcite{gangemi16b}, and compare the extended and the original profiles in a task of Word Frame Disambiguation (WFD).

\subsection{Dataset: FrameNet Full Text Documents}

To create a silver standard we processed all 108 documents from the FrameNet 1.7 dataset \cite{baker1998berkeley} with BabelFy \cite{Moroetal:14tacl}\footnote{\url{http://babelfy.org}}. By combining the original frame annotations with the automatically generated entity links we collected a total of 81,706 annotations, which we use in our experimental setting as a silver standard.

\begin{figure*}[t]
\center
%trim=left bottom right top
 \includegraphics[width=16cm]{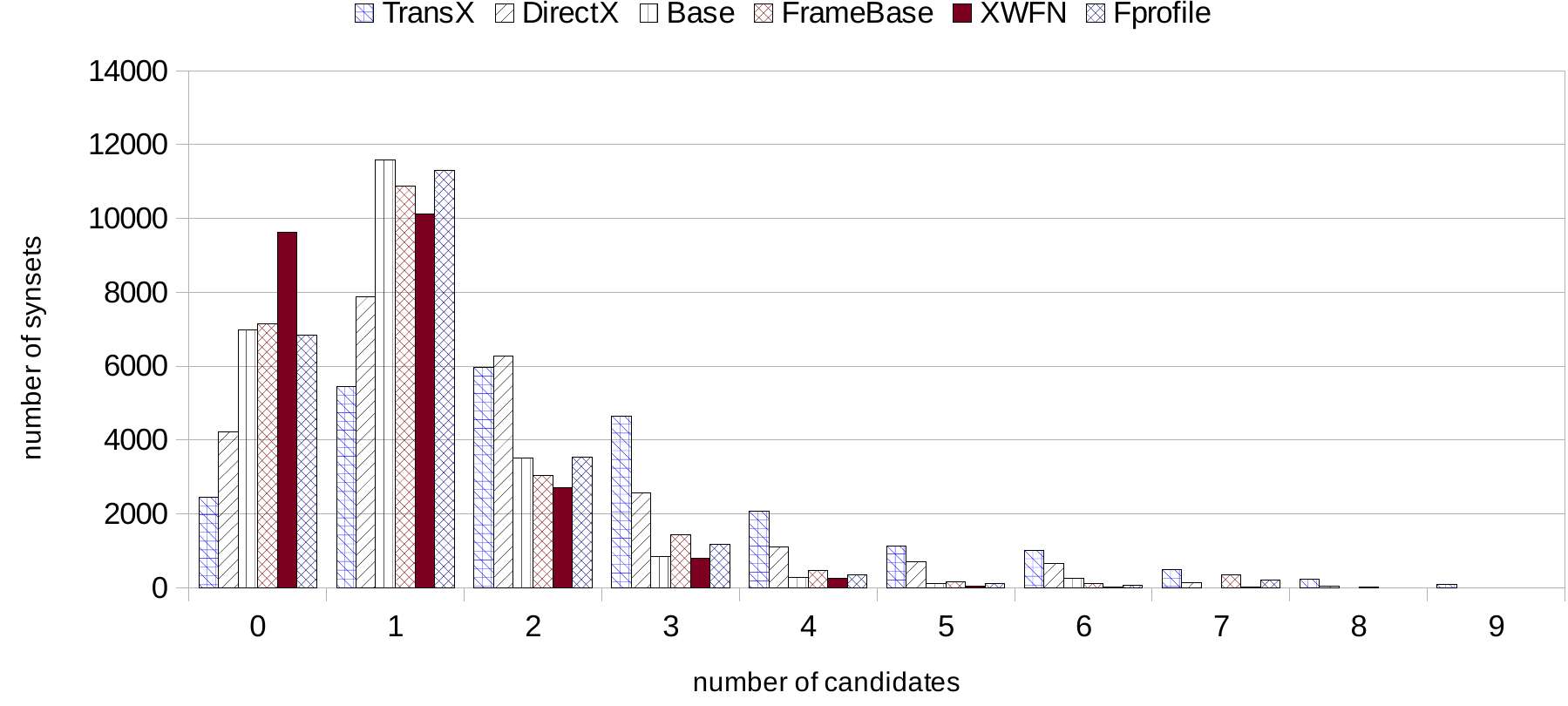}
 \caption{The distribution of synset counts per number of candidate frames (step 1 of our WFD approach, see Section \ref{sec:dis}). The y axis counts the synsets $bs$ for which $|CF(bs)|=x$. \label{fig:polisemia}}
\end{figure*}

\begin{table*}[t]
\centering
\begin{tabular}{l|rrrrrr}
      &\multicolumn{1}{c}{\textbf{Base}
}&\multicolumn{1}{c}{ \textbf{FrameBase} }   &\multicolumn{1}{c}{ \textbf{XWFN} }&\multicolumn{1}{c}{ \textbf{TransX} }&\multicolumn{1}{c}{ \textbf{DirectX} }&\multicolumn{1}{c}{ \textbf{Fprofile} }\\
\toprule
oracle &67.60      &67.74   &57.85  & 77.82  &	73.74 &	68.16 \\
\midrule
original (cond) &53.13 &	52.73 &	45.44 &	\textbf{45.21 }&	46.30 &	53.04 \\
wiki-n30-1400k (cond)&	53.23 &	52.82 &	45.42 &	44.75 &	\textbf{46.58 }&	53.46 \\
wiki-n30-1400k (inv)&\textbf{55.52 }&	\textbf{55.06 }&	\textbf{46.73 }&	28.24 &	45.39 &	\textbf{55.69 }\\

\end{tabular}
\caption{F1 performances on Word Frame Disambiguation across profiles and ranking methodologies.\label{tab:results}}

\end{table*}

\subsection{Word Frame Disambiguation}
\label{sec:dis}

Following the WFD approach described in \newcite{gangemi16b} we implemented a simple word frame disambiguator, where for each provided annotation in our silver standard we try to predict a frame label only on the basis of the BabelNet synsets generated through BabelFy. In order to provide the most suitable frame label $F$ for the provided BabelNet synset label $bs$:
\begin{enumerate}[leftmargin=4mm]
\item Creating the set of candidate frames $CF(bs)$, by collecting all the frame IDs for which $bs$ is a semantically related BabelNet synsets (e.g., $CF($\textit{bn:00085007v})$=\{$\textit{Evidence},\textit{Telling}, \textit{Communication}$\}$ see Table \ref{tab:runningloading});
\item Ranking all the candidate frames $f \in CF(bs)$ by computing the following scoring functions:
\begin{itemize}[leftmargin=3mm]
\item the score is equal to the conditional probability of a frame given a synset (\emph{cond}): 
\begin{equation*}
score_{cond}(f,bs)=\frac{w(f,bs)}{\sum_{b \in relatedSynsets(f)}{w(f,b)}};
\end{equation*}
\item the score is equal to the number of related synsets divided by the weight of the synsets (\emph{inv}): 
\begin{equation*}
score_{inv}(f,bs)=\frac{|relatedSynsets(f)|}{w(f,bs)}.
\end{equation*}
Such a scoring function promotes the candidate frame that relates with the highest number of synsets $|relatedSynsets(f)|$, and it also penalizes the selection of candidates which are triggered by synsets having higher weights, in other words promoting candidates triggered by synsets in the `long tail'; 
\end{itemize}
\item Finally, we select the candidate frame with the highest score.
\end{enumerate}

We experimented with a variety of scoring functions based on weights $w(F,bs)$ and $|relatedSynsets(F)|$, and present only the two best performing ones here. 

In Figure \ref{fig:polisemia} we draw the distribution of the counts of synsets across the number of frames that are triggered in the phase of candidate selection (step 1 of our WFD approach), i.e., in the XWFN profile, we counted ten thousand synsets $bs$ for which $|CF(bs)|=1$.

\subsection{Results}
\label{sec:res}
In Table \ref{tab:results}, we show the resulting word frame disambiguation  performances in terms of the F-score and compare across profiles and ranking methodologies. To better understand the limits of our `silver standard' and the limits of ranking methodologies we introduced a so called `oracle', which selects the labeled frame if present in the list of selected candidates independently from its position in the ranking. The results indicate that: 

\begin{itemize}[leftmargin=4mm]
\item For Base, FrameBase, XWFN and Fprofile profiles, adding distributional-semantic-based features lead to 1 to 2 points improvements of F1 in the WFD task;
\item Our approach lowers the performance of DirectX profile and even drastically lowers the performances of the TransX profile; 
\item Overall, we noticed that the former two profiles are very dense by means of related senses per frame and lead to a potential high recall (see the oracle performances on the two profiles at 73.74\%, 77.82\%  respectively) while introducing more noisy senses;
\item Finally, the presence of many frequent and noisy senses let the $score_{inv}$ function outperform other function based on weights $w(F,bs)$ and $|relatedSynsets(F)|$.  
\end{itemize}

To summarize this experiment: when using our induced weighting with the $score_{inv}$ ranking, we achieve a boost on Framester profiles that showed a good performance in the original setting, while losing performance on inferior, more noisier profiles. The overall best performance was reached by our extension of the high-quality, manually created Base profile, followed by our extension to the Fprofile. 

\subsection{Error analysis}
\label{sec:erran}
An in-depth analysis of the errors made during the experiments (see Section \ref{sec:res}) by our WFD approach (see Section \ref{sec:dis}) allowed us to identify the following categories of errors:
\begin{itemize}[leftmargin=4mm]
\item \textit{Misaligned annotations:} our silver standard is automatically created by collecting BabelFy annotations for a total of 81,706 annotations from the original 93,358 FrameNet 1.7 annotations; 
\item \textit{Entity linking errors:} we estimated that around 15\% of the 81,706 BabelFy annotations\footnote{A sample of 1000 annotations has been manually annotated as correct or wrong by two annotators, and a third annotator provided a final judgment only for the annotations in disagreement.} assigned a wrong sense, eventually triggering the selection of a wrong frame;
\item \textit{Ranking error:} errors where an inferior ranking leads to bad performance. These are a consistent part of the WFD errors and may be improved with more complex weighting/ranking approaches;
\item \textit{Profile errors:} related to a specific profile. In this category we identify errors due to: i) the slightly difference between some frame meaning (e.g., many errors are due to the prediction of the frame \textit{Measure duration} instead of \textit{Calendric unit}); ii) the frequent absence of related senses so that is not possible to trigger any frame, even for correct BabelFy annotations.
\end{itemize}

\section{Conclusions}

In this paper, we presented \textit{LOaDing}, a novel lexical resource that connects distributional sense representations induced from text to symbolic frame representations. Hybrid semantic representations have been shown to enable complex semantic tasks like, for instance, end-to-end taxonomy induction \cite{Farallietal:2017}: they could provide an additional signal to improve state-of-the-art semantic parsing (as illustrated by our running example). As we have shown for word sense disambiguation for nouns \cite{panchenko2017sense}, distributional information is able to considerably alleviate the sparsity inherent in knowledge-based methods. Consequently, in future work we would like to study the contribution and the potential of our proto-conceptualizations into more complex frame-centered tasks such as verb frame induction \cite{vulic-mrkvsic-korhonen:2017:EMNLP2017}.  Our vision in the longer term is to exploit hybrid statistical and symbolic approaches to go beyond the vocabulary and relations of entity- and encyclopedic-centric resources to produce novel semantic representations of event templates and frames with a large coverage.

\section*{Acknowledgements}
This work was funded by the Deutsche Forschungsgemeinschaft (DFG) within the JOIN-T project.

%\vspace{1em}
%\noindent
%\textbf{Acknowledgments.} We acknowledge the support of the Deutsche Forschungsgemeinschaft (DFG) under the JOIN-T project.

%\vspace{1em}

\section{Bibliographical References}
\label{main:ref}

\bibliographystyle{lrec}
\bibliography{prontobib}

\begin{thebibliography}{}

\bibitem[\protect\citename{Baker \bgroup et al.\egroup
  }1998]{baker1998berkeley}
Baker, C.~F., Fillmore, C.~J., and Lowe, J.~B.
\newblock (1998).
\newblock The {B}erkeley {F}ramenet project.
\newblock In {\em Proceedings of the 36th Annual Meeting of the Association for
  Computational Linguistics and 17th International Conference on Computational
  Linguistics-Volume 1}, pages 86--90. Association for Computational
  Linguistics.

\bibitem[\protect\citename{Biemann \bgroup et al.\egroup }2018]{biemann2018}
Biemann, C., Faralli, S., Panchenko, A., and Ponzetto, S.~P.
\newblock (2018).
\newblock A framework for enriching lexical semantic resources with
  distributional semantics.
\newblock {\em Natural Language Engineering}, pages 1--48.

\bibitem[\protect\citename{Bizer \bgroup et al.\egroup }2009]{bizer09}
Bizer, C., Lehmann, J., Kobilarov, G., Auer, S., Becker, C., Cyganiak, R., and
  Hellmann, S.
\newblock (2009).
\newblock {DBpedia} -- {A} crystallization point for the web of data.
\newblock {\em JWS}, 7(3):154--165.

\bibitem[\protect\citename{Carlson \bgroup et al.\egroup }2010]{carlson10}
Carlson, A., Betteridge, J., Kisiel, B., Settles, B., Hruschka, E.~R., and
  Mitchell, T.~M.
\newblock (2010).
\newblock Toward an architecture for never-ending language learning.
\newblock In {\em Proceedings of AAAI}, pages 1306--1313, Atlanta, GA, USA.
  AAAI Press.

\bibitem[\protect\citename{Chen \bgroup et al.\egroup }2014]{7078639}
Chen, Y.~N., Wang, W.~Y., and Rudnicky, A.~I.
\newblock (2014).
\newblock Leveraging frame semantics and distributional semantics for
  unsupervised semantic slot induction in spoken dialogue systems.
\newblock In {\em Proceedings of IEEE Spoken Language Technology Workshop
  (SLT)}, pages 584--589, South Lake Tahoe, NV, USA. IEEE.

\bibitem[\protect\citename{De~Clercq \bgroup et al.\egroup }2014]{declercq14b}
De~Clercq, O., Schuhmacher, M., Hoste, V., and Ponzetto, S.~P.
\newblock (2014).
\newblock Exploiting {FrameNet} for content-based book recommendation.
\newblock In {\em Proceedings of CBRecSys Workshop at RecSys}, pages 14--20,
  Foster City, CA, USA. ACM.

\bibitem[\protect\citename{Dietz \bgroup et al.\egroup }2017]{dietz17}
Dietz, L., Kotov, A., and Meij, E.
\newblock (2017).
\newblock Utilizing knowledge graphs in text-centric information retrieval.
\newblock In {\em Proceedings of WSDM}, pages 815--816. ACM.

\bibitem[\protect\citename{Dong \bgroup et al.\egroup }2014]{Dong14}
Dong, X., Gabrilovich, E., Heitz, G., Horn, W., Lao, N., Murphy, K., Strohmann,
  T., Sun, S., and Zhang, W.
\newblock (2014).
\newblock Knowledge vault: {A} web-scale approach to probabilistic knowledge
  fusion.
\newblock In {\em Proceedings of KDD}, pages 601--610, New York, NY, USA. ACM.

\bibitem[\protect\citename{Faralli \bgroup et al.\egroup
  }2016]{Farallietal:2016}
Faralli, S., Panchenko, A., Biemann, C., and Ponzetto, S.~P.
\newblock (2016).
\newblock Linked disambiguated distributional semantic networks.
\newblock In {\em Proceedings of ISWC}, pages 56--64, Kobe, Japan. Springer
  International.

\bibitem[\protect\citename{Faralli \bgroup et al.\egroup
  }2017]{Farallietal:2017}
Faralli, S., Panchenko, A., Biemann, C., and Ponzetto, S.~P.
\newblock (2017).
\newblock The {ContrastMedium} algorithm: Taxonomy induction from noisy
  knowledge graphs with just a few links.
\newblock In {\em Proceedings of EACL}, pages 590--600, Valencia, Spain.
  Association for Computational Linguistics.

\bibitem[\protect\citename{Fillmore}1968]{fillmore:case}
Fillmore, C.~J.
\newblock (1968).
\newblock The case for case.
\newblock In {\em Universals in Linguistic Theory}, pages 2--88. Holt, Rinehart
  and Winston, New York, NY, USA.

\bibitem[\protect\citename{Gangemi \bgroup et al.\egroup }2016a]{gangemi16}
Gangemi, A., Alam, M., Asprino, L., Presutti, V., and Recupero, D.~R.
\newblock (2016a).
\newblock Framester: {A} wide coverage linguistic linked data hub.
\newblock In {\em Proceedings of EKAW}, pages 239--254, Bologna, Italy.

\bibitem[\protect\citename{Gangemi \bgroup et al.\egroup }2016b]{gangemi16b}
Gangemi, A., Alam, M., and Presutti, V.
\newblock (2016b).
\newblock Word frame disambiguation: Evaluating linguistic linked data on frame
  detection.
\newblock In {\em Proceedings of LD4IE Workshop at ISWC}, pages 23--31, Kobe,
  Japan. Springer International.

\bibitem[\protect\citename{Moro \bgroup et al.\egroup }2014]{Moroetal:14tacl}
Moro, A., Raganato, A., and Navigli, R.
\newblock (2014).
\newblock {Entity Linking meets Word Sense Disambiguation: a Unified Approach}.
\newblock {\em Transactions of the Association for Computational Linguistics
  (TACL)}, 2:231--244.

\bibitem[\protect\citename{Navigli and Ponzetto}2012]{navigli12}
Navigli, R. and Ponzetto, S.~P.
\newblock (2012).
\newblock {B}abel{N}et: The automatic construction, evaluation and application
  of a wide-coverage multilingual semantic network.
\newblock {\em ArtInt.}, 193:217--250.

\bibitem[\protect\citename{Panchenko \bgroup et al.\egroup
  }2017]{panchenko2017sense}
Panchenko, A., Faralli, S., Ponzetto, S.~P., and Biemann, C.
\newblock (2017).
\newblock Using linked disambiguated distributional networks for word sense
  disambiguation.
\newblock In {\em Proceedings of SENSE Workshop at EACL}, pages 1--7, Valencia,
  Spain. Association for Computational Linguistics.

\bibitem[\protect\citename{Rebele \bgroup et al.\egroup }2016]{rabele16}
Rebele, T., Suchanek, F.~M., Hoffart, J., Biega, J., Kuzey, E., and Weikum, G.
\newblock (2016).
\newblock {YAGO:} {A} multilingual knowledge base from {W}ikipedia, {W}ordnet,
  and {G}eonames.
\newblock In {\em Proceedings of ISWC}, pages 177--185, Kobe, Japan. Springer
  International.

\bibitem[\protect\citename{Recupero \bgroup et al.\egroup }2015]{recupero15}
Recupero, D.~R., Presutti, V., Consoli, S., Gangemi, A., and Nuzzolese, A.~G.
\newblock (2015).
\newblock Sentilo: Frame-based sentiment analysis.
\newblock {\em Cognitive Computation}, 7(2):211--225.

\bibitem[\protect\citename{Vuli\'{c} \bgroup et al.\egroup
  }2017]{vulic-mrkvsic-korhonen:2017:EMNLP2017}
Vuli\'{c}, I., Mrk\v{s}i\'{c}, N., and Korhonen, A.
\newblock (2017).
\newblock Cross-lingual induction and transfer of verb classes based on word
  vector space specialisation.
\newblock In {\em Proceedings of EMNLP}, pages 2536--2548, Copenhagen, Denmark.
  Association for Computational Linguistics.

\end{thebibliography}

%\section{Language Resource References}
%\label{lr:ref}
%\bibliographystylelanguageresource{lrec}
%\bibliographylanguageresource{xample}

\end{document}